\pdfoutput=1
\documentclass[runningheads,a4paper,reqno]{./style/llncs}
\usepackage{epsfig}
\usepackage{graphicx}
\usepackage{subfigure}
\usepackage{amsmath}
\usepackage{amssymb}
\usepackage{array}
\usepackage{color}

\newcommand{\rpm}{\raisebox{.2ex}{$\scriptstyle\pm$}}
\newcommand{\todo}[1]{\textbf{\textcolor{red}{#1}}}

\begin{document}

\mainmatter  

\title{Learning to recognize Abnormalities in Chest X-Rays with Location-Aware Dense Networks}

\titlerunning{ }

\authorrunning{G\"undel S., Grbic S., Georgescu B., Zhou K., et al}


\author{Sebastian G\"undel\textsuperscript{1,2}, Sasa Grbic\textsuperscript{1}, Bogdan Georgescu\textsuperscript{1}, S. Kevin Zhou\textsuperscript{1}, Ludwig Ritschl\textsuperscript{3}, Andreas Maier\textsuperscript{2} and Dorin Comaniciu\textsuperscript{1}}

\institute{\textsuperscript{1} Siemens Imaging Technologies, Siemens Healthineers, Princeton NJ, USA\\
\textsuperscript{2} Pattern Recognition Lab, Friedrich-Alexander-Universit\"at, Erlangen, Germany\\
\textsuperscript{3} Siemens Healthcare, Forchheim, Germany
}

\maketitle

\begin{abstract}
Chest X-ray is the most common medical imaging exam used to assess multiple pathologies. Automated algorithms and tools have the potential to support the reading workflow, improve efficiency, and reduce reading errors. With the availability of large scale data sets, several methods have been proposed to classify pathologies on chest X-ray images. However, most methods report performance based on random image based splitting, ignoring the high probability of the same patient appearing in both training and test set. In addition, most methods fail to explicitly incorporate the spatial information of abnormalities or utilize the high resolution images. We propose a novel approach based on location aware Dense Networks (DNetLoc), whereby we incorporate both high-resolution image data and spatial information for abnormality classification. We evaluate our method on the largest data set reported in the community, containing a total of 86,876 patients and  297,541 chest X-ray images. We achieve (i) the best average AUC score for published training and test splits on the single benchmarking data set (ChestX-Ray14 \cite{wang2017chestx}), and (ii) improved AUC scores when the pathology location information is explicitly used. To foster future research we demonstrate the limitations of the current benchmarking setup \cite{wang2017chestx} and provide new reference patient-wise splits for the used data sets. This could support consistent and meaningful benchmarking of future methods on the largest publicly available data sets. 

\end{abstract}

\section{Introduction}
\label{sec:intro}
Chest X-ray is the most common medical imaging exam with over 35 million taken every year in the US alone  \cite{kamel2017utilization}. They allow for inexpensive screening of several pathologies including masses, pulmonary nodules, effusions, cardiac abnormalities and pneumothorax. Due to increasing workload pressures, many radiologists today have to read more than 100 X-ray studies daily. Therefore, automated tools trained to predict the risk of specific abnormalities given a particular X-ray image have the potential to support the reading workflow of the radiologist. Such a system could be used to enhance the confidence of the radiologist or prioritize the reading list where critical cases would be read first.
\par
Due to the recent availability of a large scale data set \cite{wang2017chestx}, several works have been proposed to automatically detect abnormalities in chest X-rays. The only peer-reviewed published work is by Wang et al. \cite{wang2017chestx} which evaluated the performance using four standard Convolutional Neural Networks (CNN) architectures (AlexNet, VGGNet, GoogLeNet and ResNet \cite{he2016deep}). The following not peer-reviewed papers can be found on arXiv. In \cite{rajpurkar2017chexnet}, a slightly modified DenseNet architecture was used. Yao et al. \cite{yao2017learning} utilized a variant of DenseNet and Long-short Term Memory Networks (LSTM) to exploit the dependencies between abnormalities. In \cite{2018arXiv180109927G}, Guan et al. proposed an attention guided CNN whereby disease specific regions are first estimated before focusing the classification task on a reduced field of view. However most of the current work on arXiv shows results by splitting the data randomly for training, validation and testing \cite{2018arXiv180109927G,yao2017learning} which is problematic as the average image count per patient for the ChestX-Ray14 \cite{wang2017chestx} data set is 3.6. Thus the same patient is likely to appear in both training and test set. Additionally there is a significant variability in the classification performance between splits due to the class imbalance, thus making performance comparisons problematic. The solely prior work containing publicly released patient-wise splits is the work by Wang et al\cite{wang2017chestx}.
\par
In this paper, we propose a location aware Dense Network (DNetLoc) to detect pathologies in chest X-ray images. We incorporate the spatial information of chest X-ray pathologies and exploit high resolution X-ray data effectively by utilizing high-resolution images during training and testing. Moreover, we benchmark our method on the largest data set reported in the community with 86,876 patients and about 297,541 images, utilizing both the ChestX-Ray14 \cite{wang2017chestx} and PLCO \cite{gohagan2000prostate} data sets. In addition we propose a new benchmarking set-up on this data set, including published patient-wise training and test splits, supporting the ability to effectively compare future algorithm performance on the largest public chest X-ray data set. We achieve the best performance reported on the existing ChestX-Ray14 benchmarking data set where both patient-wise train and test splits are published.
\section{Datasets}
\label{sec:ds}

\begin{figure}[!ht]
\begin{center}
\includegraphics[width=0.85\linewidth]{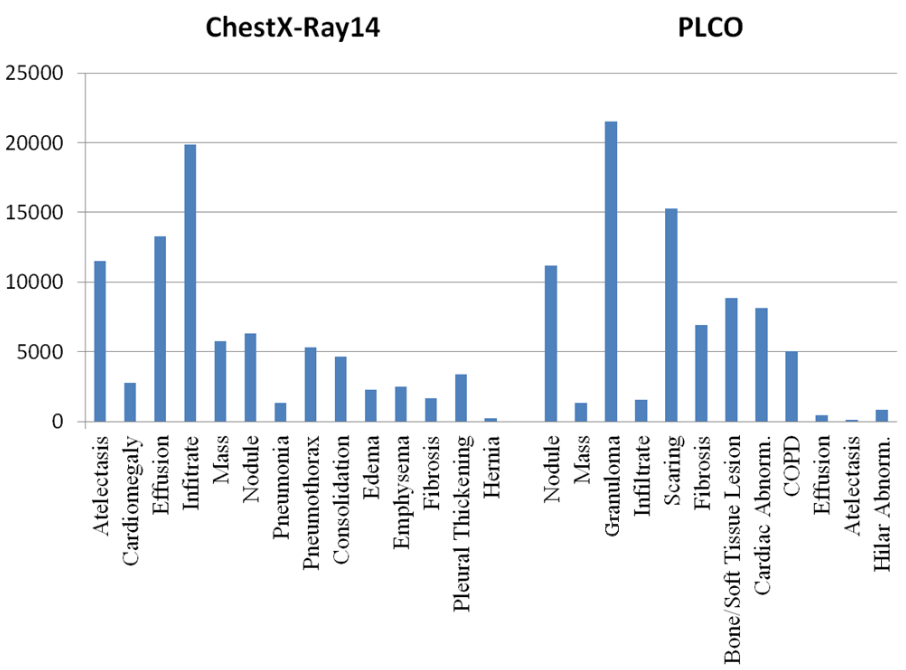}
\caption{Representation of the image numbers of the 2 datasets. The chart excludes the number of images where none of these pathologies appear.}
\label{fig:num_images}
\end{center}
\end{figure}

The ChestX-Ray14 data set \cite{wang2017chestx} contains 30,805 patients and 112,120 chest X-ray images. The size of each image is $1024 \times 1024$ with 8 bits gray-scale values. The corresponding report includes 14 pathology classes. 
\par
In the PLCO data set \cite{gohagan2000prostate}, there are 185,421 images from 56,071 patients. The original size of each image is $2500 \times 2100$ with 16 bit gray-scale values. We choose 12 most prevalent pathology labels, among which 5 pathology labels contain also the spatial information. The details of such spatial information are described in Section 3.3.
\par
Across both data sets, there are 6 labels which share the same name. However, in our experiment, we avoid combining the images of similar labels as we cannot guaratee the same label definition. Additionally we assume there is no patient overlap between these two datasets.
\par
The pathology labels are highly imbalanced. This is clearly illustrated in Fig. \ref{fig:num_images}, which displays the total number of images across all pathologies in the 2 data sets. This poses a challenge to any learning algorithm.

\section{Method}
\label{sec:method}

\subsection{Multi-label Setup}
\label{subsec:multi-label}
We use a variant of DenseNet with 121 layers \cite{huang2018archive}. Each output is normalized with a sigmoid function to [0,1]. The network is initialized with the pre-trained ImageNet model \cite{ImageNet}. At first we focus on the ChestX-Ray14 dataset. The labels consist of a \textit{C} dimensional vector $[l_1, l_2 \ldots l_C]$ where \textit{C=14} with binary values, representing either the absence (0) or the presence (1) of a pathology. As a multi-label problem, we treat all labels during the classification independently by defining \textit{C} binary cross entropy loss functions. As the data set is highly imbalanced, we incorporate additional weights within the loss functions, based on the label frequency within each batch:
\begin{equation}\label{eq:loss}
\mathcal{L}(X,l_n) = -(w_P * l_n\log(p) + w_N * (1 - l_n)log(1 - p)),
\end{equation}
where $w_P = \frac{P_n + N_n}{P_n}$ and $w_N = \frac{P_n + N_n}{N_n}$, with $P_n$ and $N_n$ indicating the number of presence and absence samples, respectively.
\par
During training, we use a batch size of 128. Larger batch sizes increase the probability to contain samples of each class and increase the weight scale of $w_P$ and $w_N$. The original images are normalized based on the ImageNet pre-trained model \cite{ImageNet} with 3 input channels. We increase the global average pooling layer before the final layer to $8 \times 8$. The Adam optimizer \cite{adam} ($\beta_1 = 0.9$, $\beta_2 = 0.999$, $\epsilon = 10^{-8}$)  is used with an adaptive learning rate: The learning rate is initialized with $10^{-3}$ and reduced 10x when the validation loss plateaus.

\subsection{Leveraging High-Resolution Images and Spatial Knowledge}
\label{subsec:downsampling}
Two strided convolutional layers with 3 filters of $3 \times 3$ and a stride of 2 are added as the first layers to effectively exploit the high-resolution chest X-ray images. The filter weights of both layers are initialized equal to a Gaussian down-sampling operation. We use an image size of $1024 \times 1024$ as input to our network.

\label{subsec:location}

\begin{figure}[ht!]
\begin{center}
\includegraphics[width=0.9\linewidth]{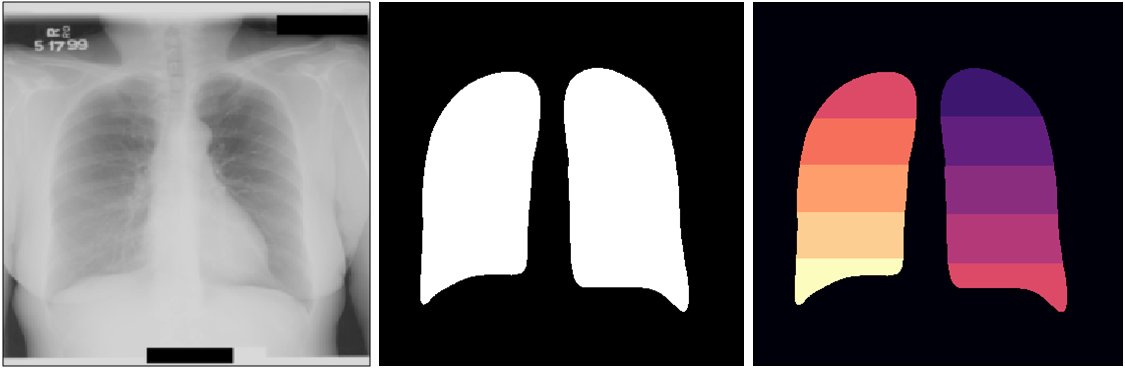}
\caption{Spatial Ground Truth Labels:  \textit{Left:} Chest X-ray image, \textit{Middle:} Lung side (information whether disease is in the left lung, right lung or neither), \textit{Right:} Splitting of the lung in fifth (more detailed information where disease is located)}
\label{fig:LabelSplit}
\end{center}
\end{figure}

Contrary to the ChestX-Ray14 \cite{wang2017chestx} data set the PLCO data set \cite{gohagan2000prostate} includes consistent spatial location labels for many pathologies. We include 12 pathology labels of the PLCO data set in our experiments (see Fig. \ref{fig:num_images}, right side). The location information is available for 5 pathologies. The location information contains the information about the side (right lung, left lung), finer localization in each lung (divided in equal fifth), including an additional label for diffuse disease. The exact position information of multiple and diffused diseases is not provided.
\par
Therefore, we create 9 additional classes: 6 are responsible for the lobe position (equally split in five parts and a ``wildcard'' label for multiple diseases: E.g. if the image contains nodules in multiple lung parts, only this label is present), 2 for the lung side (left and right), and 1 for diffused diseases over multiple lung parts. Fig. \ref{fig:LabelSplit} illustrates the label definition based on spatial information.
\par
The spatial location labels are trained as binary and independent classes with cross entropy functions. The number of present class labels depend on the number of diseases that contain location information. 

\subsection{Dataset Pooling}
\label{subsec:grouping}

We combine the ChestX-Ray14 and the PLCO datasets. The training and validation set includes images from both data sets. Several classes share similar class labels. However, we do not know if both data sets are created based on the same label definition. Due to this fact, we treat the labels independently and create different classes. We normalize brightness and contrast of the PLCO dataset images by applying histogram normalization. All images are normalized based on the mean and the standard deviation to match the ImageNet definition. Each batch contains images from both data sets.
\par



Combining both datasets (\textit{C=35}), we compute the loss function
\begin{equation}\label{eq:gl_loss}
\mathcal{L}(X) = -\frac{1}{C}\sum\limits_{n=1}^{C} w(w_P * l_n\log(p) + w_N * (1 - l_n)log(1 - p)),
\end{equation}
where $w$ is either 0 or 1, depending which dataset the image is coming from and whether the spatial information exists.

\begin{figure}[t]
\begin{center}
\includegraphics[width=0.85\linewidth]{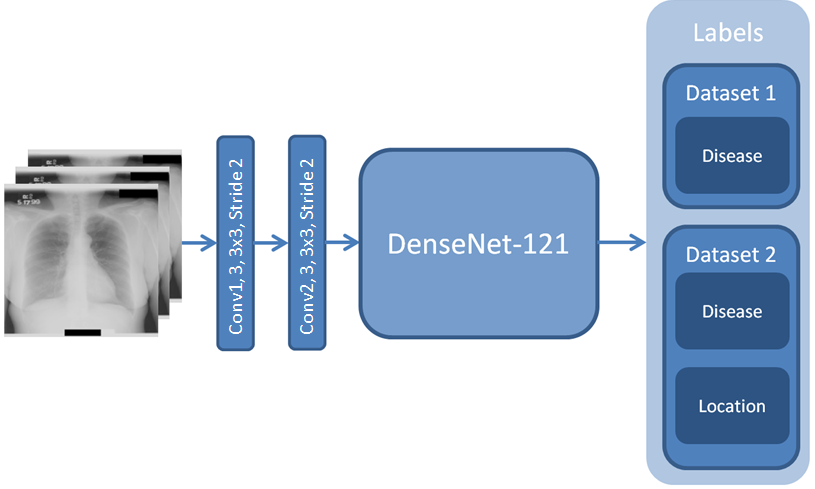}
\caption{Architecture of the proposed network (DNetLoc): The current image is provided in 3 channels, followed by 2 strided convolutional layer and the DenseNet-121. Then, the classes are triggered depending on the data set the current sample originates from. }
\label{fig:Network}
\end{center}
\end{figure}

\subsection{Global Architecture}

Overall, we create a local aware Dense Network that adaptively deals with label availability during training. The final network consists of 35 labels, 14 from the ChestX-Ray14 dataset, and 21 from PLCO dataset. Fig. \ref{fig:Network} illustrates the architecture of the network (DNetLoc). 
\section{Experimental Results}
\label{sec:experiments}
The ChestX-Ray14 dataset contains an average of 3.6 images per patient and PLCO 3.3 images per patient. Thus, there is a high probability the same patient appears in all 3 subsets if a random image-split is used. This paper uses only patient-wise splits. All splits used in this paper are published on GitHub \cite{SplitWeb}. For all experiments we separate the data as follows: 70\% for training, 10\% for validation, and 20\% for testing.
\par
Below we present our experimental results. First, we show the state-of-the-art results on the ChestX-Ray14 dataset, following the official patient-wise split. Then, we present the results on the PLCO data set, illustrating the value of using location information and data pooling.

\begin{table}[!ht]
\begin{center}
\label{tab:patient_random_split}
\begin{tabular}{| l || c | c | c |} 

\hline
Method & Wang \textit{et al.} \cite{wang2017chestx} &  Our DNet  & Our DNet\\
\hline
Official Split & Yes & Yes & No\\
\hline
\hline
Atelectasis &  0.7003 & 0.767 & 0.826\\
Cardiomegaly & 0.8100 & 0.883 & 0.911\\
Effusion &  0.7585 & 0.828 & 0.885\\
Infiltration &  0.6614 &  0.709 & 0.716\\
Mass &  0.6933 & 0.821 & 0.854\\
Nodule &  0.6687 & 0.758 & 0.774\\
Pneumonia &  0.6580 & 0.731 & 0.765\\
Pneumothorax &  0.7993 & 0.846 & 0.872\\
Consolidation &  0.7032 & 0.745 & 0.806\\
Edema &  0.8052 &  0.835 & 0.892\\
Emphysema &  0.8330 & 0.895 & 0.925\\
Fibrosis  &  0.7859 & 0.818 & 0.820\\
Pleural Thick. & 0.6835 & 0.761 & 0.785\\
Hernia &  0.8717 & 0.896 & 0.941\\
\hline
Mean &  0.7451 & 0.807 & 0.841\\
\hline
\end{tabular}
\\
\end{center}
\caption{We demonstrate improved AUC scores using our method on the official ChestX-Ray14 test set. Right column shows our method on a more representative random patient split where the mean AUC score increases to 0.841.}
\label{tab:results1}
\vspace{-0.15in}
\end{table}

\par
\begin{figure}[t]
\begin{center}
\includegraphics[width=0.9\linewidth]{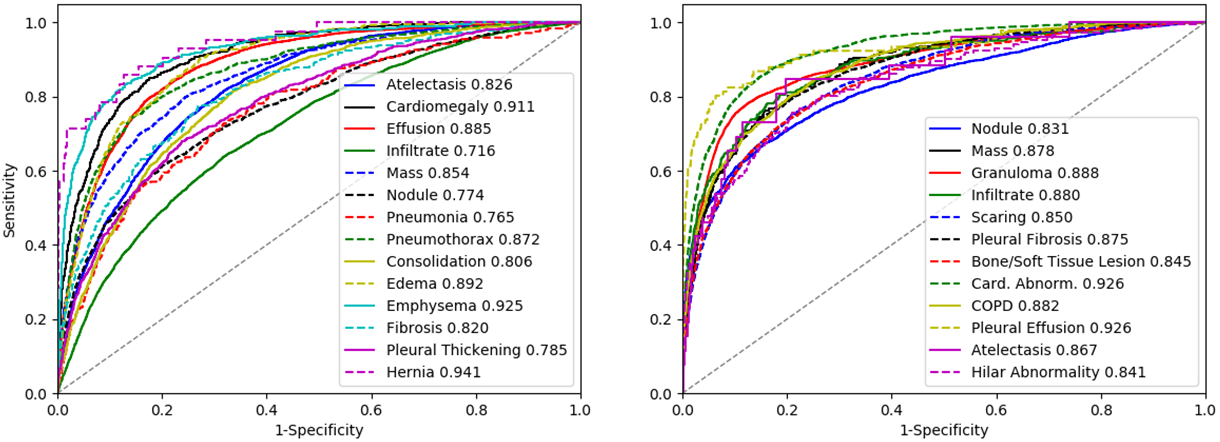}
\caption{The Area under Curve (AUC) scores for chest X-ray pathologies: \textit{Left:} Results corresponding to Table \ref{tab:results1} righ column on the ChestX-Ray14 dataset using DNet, \textit{Right:} Results corresponding to Table \ref{tab:results2} righ column on the PLCO dataset using DNetLoc.}
\label{fig:AUCscores}
\end{center}
\vspace{-0.1in}
\end{figure}

Table \ref{tab:results1} shows the best AUC scores obtained on the ChestX-Ray dataset using the official test set. Our network increases the mean AUC score by over 5\% compared to the previous work. We observed several limitations with the official split where training and test data sets have different characteristics. This can be either the large label inconsistency or the fact that there are on average 3 times more images per patient in the test set compared with the training set. Thus we computed several random patient-splits each leading better performance with average $0.831$ AUC with $0.019$ standard deviation. Detailed performance for the novel benchmarking patient-wise split is shown in Table \ref{tab:results1} and in Fig. \ref{fig:AUCscores} left).
\par
Overall, significant label variance of the follow-up exams are noticeable across the ChestX-Ray14 data set. This might be due to the circumstance that many follow-ups are generated with a specific question, e.g did the Pneumothorax disappear. Thus repeated and consistant labeling of other abnormalities in follow-up studies varies. As the ChestX-Ray14 labels are generated from reports this would introduce incomplete labeling for many follow-ups.

\par

\begin{table}[t]
\begin{center}
\begin{tabular}{| l || c | c |} 

\hline
Method & Our DNet & Our DNetLoc\\
\hline
\hline
\textbf{Nodule} & \textbf{0.817} & \textbf{0.831} \\
\textbf{Mass} & \textbf{0.845} & \textbf{0.878}\\
Granuloma & 0.888 &  0.888\\
\textbf{Infiltrate} & \textbf{0.875} &  \textbf{0.880}\\
Scaring & 0.841 &  0.850\\
Fibrosis & 0.873 & 0.875\\
Bone/Soft Tissue Lesion & 0.853 &  0.845\\
Cardiac Abnormality & 0.927 & 0.926\\
COPD & 0.881 &  0.882\\
Effusion & 0.933 & 0.926\\
\textbf{Atelectasis} & \textbf{0.831} & \textbf{0.867}\\
\textbf{Hilar Abnormality}  & \textbf{0.812} &  \textbf{0.841}\\
\hline
\textbf{Mean (Location)} &  \textbf{0.836} & \textbf{0.859}\\
\hline
Mean & 0.865 & 0.874\\
\hline

\end{tabular}
\\
\end{center}
\caption{Test results on the PLCO data set: DNetLoc achieves the best AUC scores on the PLCO dataset. Both DNet and DNetLoc networks were trained on the combined  ChestX-Ray14 and PLCO dataset. DNetLoc uses spatial information of 5 pathologies. The bold values in this table are pathologies which are supported by spatial knowledge.}
\label{tab:results2}
\vspace{-0.15in}
\end{table}

\par
Finally we evaluate our method on the PLCO data. The results in Table \ref{tab:results2} show that location information and leveraging high resolution images improve the classification accuracy for most pathologies. For a subset of  pathologies where location information is provided (marked in bold), the performance increases by an average of 2.3\%. Moreover, the training time was reduced by a factor of 2 when location information is used. For the PLCO data set we reach a final mean AUC score of 87.4\%. Fig. \ref{fig:AUCscores} shows the performance of our method for both the ChestX-Ray14 and the PLCO test set.
\par


\section{Conclusion}
\label{sec:Conclusion}
We presented a novel method based on location aware Dense Networks to classify pathologies in chest X-ray images, effectively exploiting high-resolution data and incorporating spatial information of pathologies to improve the classification accuracy. We showed that for pathologies where the location information is present the classification accuracy improved significantly. The algorithm is trained and validated on the largest chest X-ray data set containing 86,876 patients and 297,541 images. Our system has the potential to support the current high throughput reading workflow of the radiologist by enabling him to gain more confidence by asking an AI system for a second opinion or flag ''critical'' patients for closer examination. In addition we have shown the limitations in the validation strategy of previous works and propose a novel setup using the largest public data set and provide patient-wise splits which will facilitate a principled benchmark for future methods in the space of abnormality detection on chest X-ray imaging.


\textbf{Disclaimer}: This feature is based on research, and is not commercially available. Due to regulatory reasons, its future availability cannot be guaranteed.

\bibliographystyle{./style/splncs}
\bibliography{MICCAI2018XRay}
\end{document}